\documentclass[letterpaper, 10 pt, conference]{ieeeconf}  

\IEEEoverridecommandlockouts                              

\usepackage{graphics} 
\usepackage{epsfig} 
\usepackage{mathptmx} 
\usepackage{times} 
\usepackage{amsmath} 
\usepackage{amssymb}  
\usepackage{xcolor,colortbl}
\usepackage{subfigure}
\usepackage{hyperref}
\usepackage{bbm}
\usepackage{bm}
\usepackage{algorithm, algpseudocode}

\usepackage[nolist]{acronym}
\usepackage[textsize=tiny]{todonotes}
\usepackage{placeins}
\usepackage{bigstrut}
\usepackage{gensymb}
\usepackage[font=footnotesize]{caption}
\usepackage{booktabs}

\DeclareMathOperator*{\argmax}{argmax}

\DeclareMathOperator*{\E}{\mathbb{E}}

\newcommand{\ve}{\mathbf{e}}
\newcommand{\vg}{\mathbf{g}}

\newcommand{\vo}{\mathbf{o}}
\newcommand{\vx}{\mathbf{x}}
\newcommand{\va}{\mathbf{a}}

\newcommand{\vtau}{\mathbf{\tau}}
\newcommand{\mR}{\mathbf{R}}
\newcommand{\mM}{\mathbf{M}}
\newcommand{\mT}{\mathbf{T}}
\newcommand{\mI}{\mathbf{I}}
\newcommand{\vtheta}{\mathbf{\theta}}
\newcommand{\mSig}{\bm{\Sigma}}
\newcommand{\mPhi}{\bm{\Phi}}

\newcommand{\vsig}{\bm{\sigma}}

\newcommand{\mK}{\mathbf{K}}

\DeclareMathAlphabet{\mathpzc}{OT1}{pzc}{m}{it}
\DeclareMathAlphabet{\mathcal}{OMS}{cmsy}{m}{n}
\DeclareMathAlphabet{\mathbfcal}{OMS}{cmsy}{b}{n}

\newcommand{\Ds}{\mathcal{D}}

\long\def\invis#1{}

\newcommand\gderror[1]{
   \typeout{--------------------------------------------------------------------}
   \typeout{------- #1 ---------}
   \typeout{--------------------------------------------------------------------}
   {\bf #1}
}
\newcounter{gdTmp}
\newcounter{gdLastCount}
\newcommand\maxpage[2][Error]{  
\ifnum\value{page}>#2
    \gderror{{\Large On page {\thepage} we are past page #2 (too long).   #1 }}
\else\fi
\setcounter{gdLastCount}{\value{page}} 
}
\newcommand\maxpageSinceLast[2][Error]{  
\ifnum \numexpr \value{page} - \value{gdLastCount}\relax>#2
    \gderror{Exceeds max length #2 pages. Page \thepage: #1}
\thepage\else\fi
\setcounter{gdLastCount}{\value{page}}
}

\title{\LARGE \bf
Trajectory-Constrained Deep Latent Visual Attention for Improved Local Planning in Presence of Heterogeneous Terrain
}

\author{Stefan Wapnick, Travis Manderson, David Meger, Gregory Dudek
\thanks{Mobile Robotics Laboratory, 
	School of Computer Science,
	McGill University, Montreal, Quebec, Canada\newline
        {\tt\small \{swapnick,travism,dmeger,dudek\}@cim.mcgill.ca}}%
}

\newcommand{\norm}[1]{\left\lVert#1\right\rVert}
\newcommand{\etal}{\textit{et al.}}

\begin{document}

\setlength{\textfloatsep}{10pt}

\bstctlcite{MyBSTcontrol}


\maketitle
\thispagestyle{empty}
\pagestyle{empty}

\begin{abstract}

We present a reward-predictive, model-based deep learning method featuring trajectory-constrained visual attention for local planning in visual navigation tasks. Our method learns to place visual attention at locations in latent image space which follow trajectories caused by vehicle control actions to enhance predictive accuracy during planning. The attention model is jointly optimized by the task-specific loss and an additional trajectory-constraint loss, allowing adaptability yet encouraging a regularized structure for improved generalization and reliability. Importantly, visual attention is applied in latent feature map space instead of raw image space to promote efficient planning. We validated our model in visual navigation tasks of planning low turbulence, collision-free trajectories in off-road settings and hill climbing with locking differentials in the presence of slippery terrain. Experiments involved randomized procedural generated simulation and real-world environments. We found our method improved generalization and learning efficiency when compared to no-attention and self-attention alternatives.
\end{abstract}


\section{Introduction}
\label{sec:intro}

In this paper we present a model-based, reward-predictive method augmented with trajectory-constrained visual attention to improve predictive accuracy during local planning over control actions in visual navigation tasks.

The deployment of autonomous vehicles in unstructured natural environments has a variety of applications such as environmental monitoring and crisis management. Small-scale vehicles are desirable in such settings due to their reduced environmental impact, ability to navigate in narrow access spaces, and relatively inexpensive hardware. However, navigation in unstructured natural environments possesses its own set of challenges. These includes diverse terrain, obstacle, and foliage configurations of variable traversability, changing weather conditions, and the resulting fluctuating effects on vehicle dynamics and sensor inputs.

Navigation in traditional robotics involves steps of mapping, localization, and path planning using dynamics models of assumed structure and cost maps derived from the environment. However, the complexity of unstructured natural environments can make explicit cost mapping and dynamics modeling under imposed structure brittle due to the immense variety and unpredictability of possible states encountered.

Recently, data-driven deep learning methods have seen success in creating navigation and control policies in robotics \cite{Giusti2016, Smolyanskiy2017, bojarski2016end, Wellhausen2019, kahn2020badgr, air_ground_planning}. Deep learning methods learn a function mapping minimizing a cost metric evaluated on data while requiring limited explicit user modeling and offering the prospect of continual improvement. Yet, deep learning methods are criticised for their lack of transparency~\cite{iyer_transparency_2018}, difficulty generalizing to inputs different from the training dataset~\cite{NIPS2018_outofdistribution}, and large data requirements~\cite{deep_learning_book}. This is problematic in applied robotics where safety requires model consistency and gathering real data may be difficult.
Furthermore, although visual self-attention has been known to improve performance \cite{jetley2018_learn_to_pay_attention, Schlemper2019_attentiongatednetworks, vaswani2017_transformers}, it is often unconstrained and thus potentially unreliable in what traits are learned.

\begin{figure}[t]
  \centering
  \includegraphics[width=1.0\linewidth]{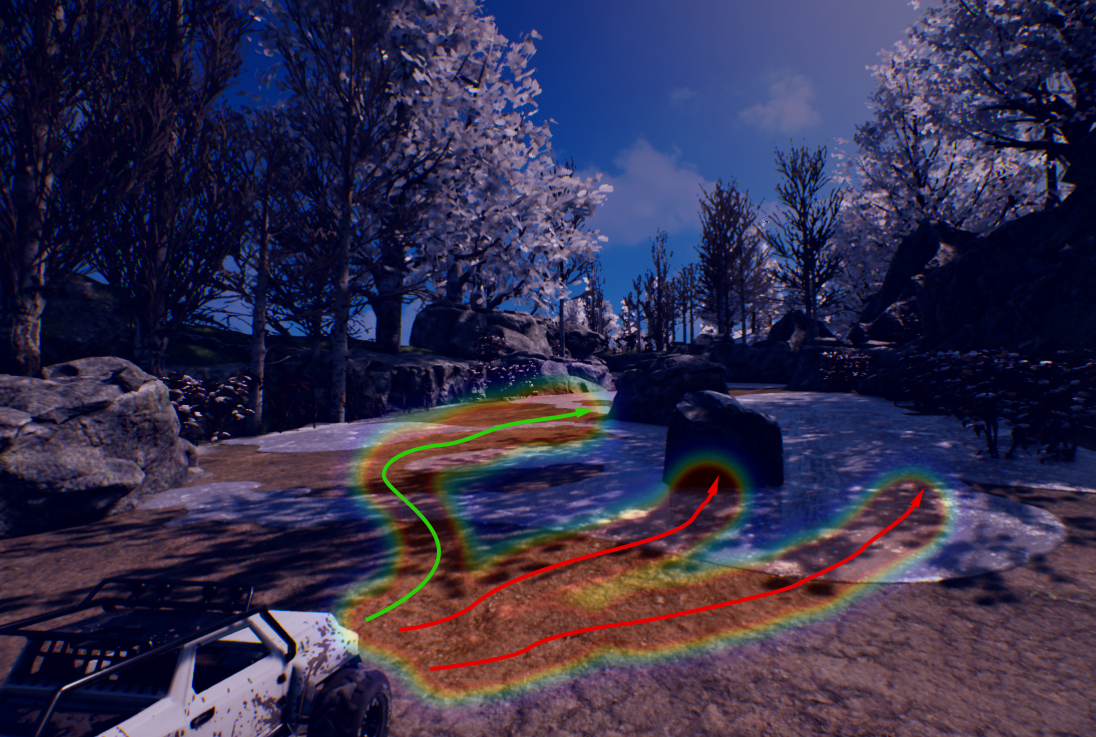}
  \caption{Our method learns visual attention locations in latent space (seen projected back into image space here) which align with trajectories implied by vehicle control actions and then uses this learned attention model to better predict candidate trajectory rewards during planning.}
\label{fig:overview}
\end{figure}

Our learning method addresses issues of generalization, reliability and data efficiency in local visual navigation tasks by applying trajectory-constrained visual attention as illustrated in Fig.~\ref{fig:overview}. During training, our model learns to associate control actions of trajectories to visual attention locations in image latent feature map space. This learned attention model is later used to focus candidate trajectories during action-space planning and enhance the predictive accuracy of each trajectory's worth. An additional guiding trajectory-constraint loss is used to enforce a reliable attention mechanism even when faced with new inputs. Filtering by this attention mask removes unnecessary background features which may be correlated with a trajectory's data labels, but are not causal, and so prevents unexpected overfitting while also producing a lower dimensional yet high information representation for improved learning efficiency.

\section{Related Work}
\label{sec:related}

\subsection{Deep Learning Methods for Navigation Control Policies}


\textbf{Imitation Learning:}
Imitation learning is a supervised method which learns a control policy from demonstrated expert actions. Pomerleau~\etal~\cite{pomerleau1989alvinn} presented one of the first applications of imitation learning for road following involving neural networks. However, direct imitation learning policies are not robust to when the learner drifts from the demonstrated state distribution. Several approaches augment training data with samples of failure cases and corrective actions to counteract learner-expert drift \cite{Giusti2016, Smolyanskiy2017, bojarski2016end}. The DAgger algorithm~\cite{ross2011reduction} and its variants involving uncertainty modeling~\cite{ensemble_dagger_2019, hg_dagger_2019} iteratively query the expert through multiple learner deployments. Other imitation learning methods use forms of sensor fusion as a fallback in case of drift \cite{lee_sensor_fusion_uncertainty_2019}. Nonetheless, the principal disadvantage of imitation learning methods is the requirement for expert demonstration.

\textbf{Reinforcement Learning:}
Reinforcement learning methods seek to learn a control policy that maximizes the expected cumulative reward over an episode using feedback from the agent's direct interaction with the environment. Reinforcement learning is divided into model-free and model-based methods. Model-based reinforcement learning is generally more sample efficient and thus commonly employed in robotics with limited data. In model-based reinforcement learning, the agent first learns a transition model of the environment which acts as a proxy for evaluating simulated state-action trajectory rollouts during planning. 
Some more recent methods learn a model for latent space dynamics and reward prediction instead of planning over raw state space~\cite{Oh2017, hafner2018planet, hafner2020dreamer, hafner2021mastering}.

Specifically, our work augments lightweight real-time reward-predictive models applied in field robotics by Khan~\etal~\cite{kahn2020badgr, Kahn2018} and Manderson~\etal~\cite{air_ground_planning} with trajectory-constrained latent visual attention to improve performance.

\subsection{Visual Attention}

Visual attention weighs areas of the observation image or corresponding convolutional feature maps based on a notion of importance and is divided into hard and soft categories. Hard attention makes discrete selections (often a location or subregion with binary weighting) while soft attention usually considers all locations with continuous weighting. 

Mnih~\etal~\cite{RAM_VisualAttention} applied multi-scale attention in image space to improve object classification accuracy using the notion of a glimpse sensor. Xu~\etal~\cite{pmlr-v37-xuc15} examined hard and soft attention in feature map space for image captioning where a softmax function was used for probability normalization and also discouraged selection of unnecessary regions as selecting larger areas results in lower normalized mask weights overall. Similar normalization and visual attention techniques have been applied to tasks of object detection and semantic segmentation \cite{jetley2018_learn_to_pay_attention, Schlemper2019_attentiongatednetworks}. The self-attention transformer architecture~\cite{vaswani2017_transformers} has also been used in several vision tasks such as image generation~\cite{pmlr-v80-parmar18a_image_transformer} and object detection~\cite{carion2020endtoendtransformers}.

Visual attention has seen use in navigation tasks as well. Lee~\etal~\cite{lee2019_perceptualattncontrol} trained a two-step model involving trajectory spline prediction followed by image cut-outs along this spline to improve imitation learning. Drews~\cite{drews2019} made use of visual attention and a convolutional recurrent network to extract image costmaps later used by a \ac{MPC} system. Zhang~\etal~\cite{Zhang_2018_ECCV} and Liu~\etal~\cite{liu2021gaze} investigated attention from human eye gaze to enhance imitation learning in visuomotor tasks.

\begin{figure*}[t!]
  \centering
  \includegraphics[width=0.89\linewidth]{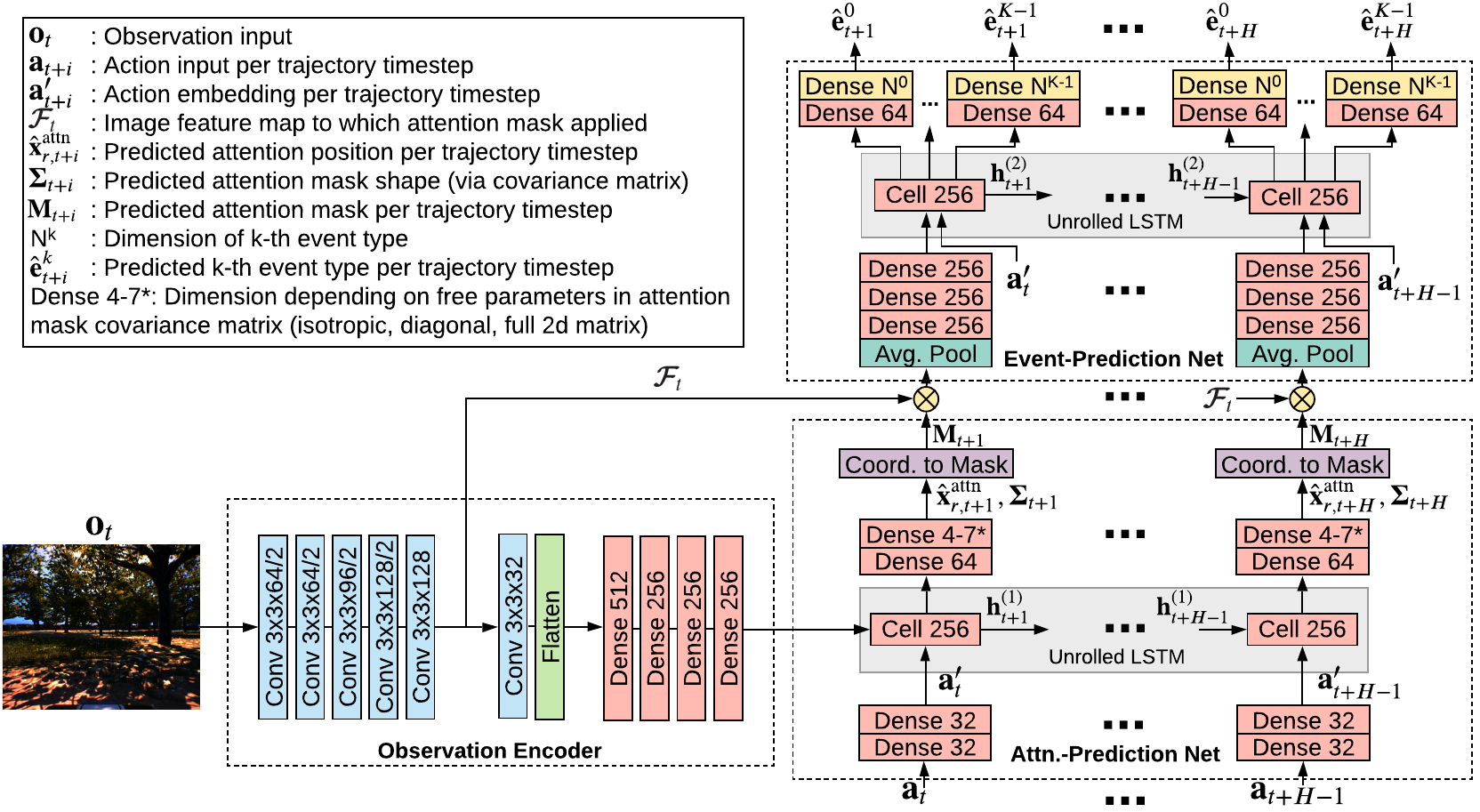}
  \caption{Proposed network made of a convolutional encoder and two stacked recurrent networks (shown unrolled through time with shared weights) to compute attention masks and events (from which rewards are derived) at each timestep over a trajectory of length $H$. Attention masks are multiplied against convolutional feature map $\mathbfcal{F}_t$ (broadcast across channels) before being input into the final event-predicting network. Layers use ReLU activation by default. Predictive outputs use softmax or no activation function depending on if discrete or continuous quantities. Convolutions are written as $k \times k \times c/s$ where $k$ is the kernel size, $c$ is the channel size and $s$ is the stride (1 if not shown). All convolutions use padding and so only strided convolutions reduce feature map width and height. Dense layers indicate the number of neurons. Average pooling is done globally. The \textbf{\textit{Coord. to Mask}} operation is described in \ref{sec:approach_attention_mask_creation}.}
\label{fig:network_architecture}
\end{figure*}

\section{Approach}
\label{sec:approach}

Our approach uses a two-stage process of model learning and planning. Section~\ref{sec:approach_representation} reviews the event or reward-predictive representation of model learning and planning (as seen in ~\cite{kahn2020badgr, Kahn2018}) while sections~\ref{sec:approach_network_architecture},~\ref{sec:approach_attention_mask_creation}, and~\ref{sec:approach_model_optimization} discuss the novel proposals for network architecture, attention generation, and optimization for implementing our notion of trajectory-constrained visual attention in this framework.

\subsection{Representation}
\label{sec:approach_representation}

Inspired by \cite{kahn2020badgr, Kahn2018}, our network learns a mapping between the current observation state $\vo_{t}$ and action sequence $\va_{t:t+H-1}$ (where $:$ denotes an inclusive index range) to future events of $K$ types over a trajectory of $H$ timesteps $\ve_{t+1:t+H}^{0:K-1}$. Specifically, we learn the function $f: \vo_{t}, \va_{t:t+H-1} \rightarrow \ve_{t+1:t+H}^{0:K-1}$. Events at each trajectory timestep represent information such as collisions, terrain turbulence or change in position which are transformed into a reward signal during planning. When change in position is estimated it should be stressed that this method does not substitute for robust localization, but is instead only used in a future predictive fashion to navigate towards goal locations in the context of waypoint following. 

This approach can be viewed as simplified model-based reinforcement learning where the policy optimization horizon is truncated to a fixed short-term length of $H$ timesteps and hence only functions as a local planner. Following the two-step model-based reinforcement learning procedure, the method is divided into model learning and planning phases. 

During model learning, we seek to maximize the log-likelihood of observed events from training trajectories of $H$ timesteps. We assume events are conditionally independent given the inputs, yielding the objective function: 
\begin{align}\label{eq:prob_representation}
\vtheta^{*} & = 
\argmax_{\vtheta}{
\sum_{\vtau_{t} \in \Ds}{
\sum_{k=0}^{K-1}{
\sum_{i=1}^{H}{\log{p(\ve_{t+i}^{k}=\tilde{\ve}_{t+i}^{k}\mid \va_{t:t+i-1}, \vo_t, \vtheta)}}
}}}
\end{align}
where $\Ds$ is the training dataset, $\vtau_{t}= (\vo_t, \va_{t:t+H-1}, \tilde{\ve}_{t+1:t+H}^{0:K-1})$ is a training trajectory sample starting at timestep $t$, $\vtheta$ are the model weights and $\tilde{\ve}_{t+i}^{k}$ denotes the specific observed label.

During planning, predicted trajectory events $\hat{\ve}_{t+1:t+H}^{0:K-1}$, where $\hat{\ve}_{t+i}^k=\E_{p(\ve_{t+i}^{k}\mid \va_{t:t+i-1}, \vo_t, \vtheta)}[\ve_{t+i}^k]$ is each event's predicted expected value learned by the network, are mapped to a trajectory reward $R$ using a task-specific function:
\begin{equation}\label{eq:reward_definition}
R(\hat{\ve}_{t+1:t+H}^{0:K-1}) = \sum_{i=1}^{H}{f_{\mathrm{reward}}({\hat{\ve}}^{0:K-1}_{t+i})}
\end{equation}
Using this reward-predictive model, we optimize for an action rollout that maximizes the predicted reward over the planning horizon of $H$ timesteps.
\begin{equation}\label{eq:planning_optimization}
\va_{t:t+H-1}^* = \argmax_{\va_{t:t+H-1}}{R(\hat{\ve}_{t+1:t+H}^{0:K-1})}
\end{equation}
The first action of this rollout is used as the agent's next command, and planning is repeated every timestep.

\subsection{Network Architecture}
\label{sec:approach_network_architecture}

Fig.~\ref{fig:network_architecture} shows the proposed network architecture modeling the mapping $f: \vo_{t}, \va_{t:t+H-1} \rightarrow \ve_{t+1:t+H}^{0:K-1}$ over a trajectory of $H$ timesteps which augments the baseline architecture in~\cite{kahn2020badgr} with trajectory-constrained visual attention. The current image observation $\vo_t$ is put through a convolutional neural network to form the initial hidden state of the lower \ac{LSTM}~\cite{Hochreiter1997} recurrent network. This recurrent network operates on the image and action embeddings to output attention positions which are in turn converted to attention masks at each timestep. Section~\ref{sec:approach_attention_mask_creation} discusses the \textbf{\textit{Coord. To Mask}} operation in more detail. Attention masks are applied against image feature map $\mathbfcal{F}_t$ and the filtered feature maps per timestep are used for final event predictions by the second upper \ac{LSTM}.

We apply attention in latent feature map space instead of raw image space to enforce real-time planning. Attention masks are a function of the input action sequence, and as described in \ref{sec:approach_planning}, planning involves evaluation of many candidate action rollouts. By applying attention in latent space, we compute the input image embedding once and re-use its value for all simulated action rollouts during planning.

\subsection{Trajectory-Constrained Attention Mask Creation}
\label{sec:approach_attention_mask_creation}

To encourage predicted attention masks to follow trajectories implied by the agent's control actions, we add an additional loss term during training which minimizes the distance between the predicted attention mask positions and observed vehicle trajectory positions (section~\ref{sec:approach_model_optimization}). This requires a state estimator to collect vehicle poses during training, but after which our model can estimate attention positions for novel inputs independently and the state estimator can be removed. Specifically, during training we learn a mapping $f^\mathrm{attn}: \vo_t, \va_{t:t+H-1} \rightarrow \vx^\mathrm{attn}_{t+1:t+H}$ where $\vx^\mathrm{attn}_{t+i}$ is the attention position in feature map space at each trajectory timestep.

In practice we find best results when the network initially predicts attention positions in the robot's coordinate frame $\vx^\mathrm{attn}_{r}$ instead of directly in 2d feature map space $\vx^\mathrm{attn}$, as the latter has reduced resolution and is scaled by changing inverse depth values while the per timestep change in robot position is more predictable. The predicted $\vx^\mathrm{attn}_{r}$ is then projected to $\vx^\mathrm{attn}$ in feature map space. 

Training data in world coordinates ($\vx^\mathrm{attn}_{w}$) are put in the robot's coordinate frame ($\vx^\mathrm{attn}_{r}$) while conversion to camera ($\vx^{attn}_{c}$) and pixel ($\vx^{attn}_{p}$) coordinates are done by embedding the projection step as a differentiable operator in the network. 
\begin{align}
\vx^\mathrm{attn}_{r} & = \mR_{w\rightarrow{}r}\mT_{w\rightarrow{}r}\vx^\mathrm{attn}_{w}\label{eq:coord_projection}\\
\vx^\mathrm{attn}_{p} & = \frac{1}{d}\mK\mR_{r\rightarrow{}c} \mT_{r\rightarrow{}c}\vx^\mathrm{attn}_{r} \label{eq:coord_projection_px}
\end{align}
where $\mR_{a\rightarrow{}b}$, $\mT_{a\rightarrow{}b}$ are rotation and translation matrix operators from frame $a$ to $b$, $d$ is the depth coordinate for normalization, and $\mK$ is the intrinsic camera matrix.

To calculate the attention position in the $L$-th feature map layer of the convolutional network, we select the feature map position whose receptive field centers the $\vx^\mathrm{attn}_p$ pixel coordinate. For the simple case where each convolution possesses padding $2p=k-1$ (where $p$ is padding and $k$ is kernel size), this position can be found by simply dividing by the net convolutional output stride such that $\vx^\mathrm{attn} = \vx^\mathrm{attn}_{p}/s_\mathrm{out}$, where $s_\mathrm{out} = \prod_{l=1}^Ls_{l}$ and $s_l$ is the stride of the $l$-th layer.

Finally, the predicted feature map coordinate is converted to a two-dimensional mask $\mM$ where each $m_{i,j}$ value is given by evaluating the scaled Gaussian:
\begin{equation}\label{eq:mask_gaussian}
m_{i, j} = |\mSig|^{-\frac{1}{2}}\mathrm{exp}(-\frac{1}{2}(\vx^\mathrm{mask} - \vx^\mathrm{attn})^T\mSig^{-1}(\vx^\mathrm{mask} - \vx^\mathrm{attn}))
\end{equation}
where $\vx^\mathrm{mask} = (i, j)$ is each $i$, $j$ location in the mask and $\vx^\mathrm{attn} = (x^\mathrm{attn}, y^\mathrm{attn})$ is the predicted attention position.

We also let the network predict the mask shape by learning $\mSig$. 
During experiments we use a simple isotropic Gaussian $\mSig=\sigma^2\mI$ where the network predicts $\log(\sigma^2)$ to force positive values. We tested diagonal $\mSig=\mathrm{diag}(\vsig^2)$ and full 2d matrix (we predict $\mPhi$, where $\mSig=\mPhi\mPhi^T+\epsilon\mI$, $\epsilon>0$ to make positive definite) forms as well, but did not find noticeable gains.

\subsection{Model Optimization}
\label{sec:approach_model_optimization}

The maximum likelihood objective of Eq.~\ref{eq:prob_representation} is converted to a concrete loss over training trajectory samples (having empirical dataset distribution described by $p_{\Ds}$) by considering that observed events may either be discrete or continuous quantities, resulting in a cross-entropy loss $L_\mathrm{disc.task}$ or squared error loss $L_\mathrm{cont.task}$ (since a Gaussian distribution is assumed in this case) respectively. A final loss term $L_\mathrm{attn}$ is added to constrain predicted attention positions to follow vehicle trajectories, as discussed in section~\ref{sec:approach_attention_mask_creation}.
\begin{align}\label{eq:loss_total}
L_\mathrm{total} & = \E_{\vtau_{t}\sim p_{\Ds}}{[L_\mathrm{cont. task} + L_\mathrm{disc. task} + L_\mathrm{attn}]}\\
L_\mathrm{cont.task} & = \sum_{k\in K_\mathrm{cont.}}{
\sum_{i=1}^{H}{
    \norm{\hat{\ve}_{t+i}^{k} - \tilde{\ve}_{t+i}^{k}}_{2}^{2}
}}\\
L_\mathrm{disc.task} & = \sum_{k\in K_\mathrm{disc.}}{
\sum_{i=1}^{H}{
    -\E_{\ve_{t+i}^{k}\sim p^{*}}{ \log{p(\ve_{t+i}^{k}\mid \va_{t:t+i-1}, \vo_t, \vtheta)} }
}}\\
L_\mathrm{attn} & = \sum_{i=1}^{H} \norm{\hat{\vx}_{r, t+i}^\mathrm{attn} - \tilde{\vx}_{r, t+i}^\mathrm{attn}}_{2}^{2}\label{eq:loss_attn}
\end{align}
where $\hat{\ve}, \hat{\vx}$ and $\tilde{\ve}, \tilde{\vx}$ denote predicted and observed values while $p^*$ is the target distribution for classification loss.
\subsection{Planning}
\label{sec:approach_planning}

With the learned reward-predictive model, we search for an action rollout that maximizes the expected trajectory reward in Eq.~\ref{eq:planning_optimization} using the \ac{CEM}~\cite{cem_method}. This iterative optimization method involves the repeated evaluation of $N$ simulated action rollouts drawn from a Gaussian distribution where each iteration updates a running mean and variance of the distribution using the top scoring $N_\mathrm{top}$ candidate action sequences. The first action of the maximizing $\va^*_{t:t+H-1}$ rollout is then used as the agent's next control action, and planning is repeated each timestep.

\section{Experiments}


\subsection{Baselines}
\label{sec:experiment_baselines}

We compared against baseline no-attention (using the architecture of~\cite{kahn2020badgr}) and self-attention models. The no-attention model uses the network in Fig.~\ref{fig:network_architecture}, but with the attention prediction module removed such that the embeddings are directly input to the final event-predicting recurrent network. The self-attention variant also uses the network in Fig.~\ref{fig:network_architecture}, but with no trajectory constraint loss or imposed attention mask structure. The \textbf{\textit{Coord. to Mask}} function is removed and the attention prediction network directly outputs a 2d mask at each timestep learned purely from the task-specific loss. A softmax activation is added (similar to ~\cite{jetley2018_learn_to_pay_attention}) to discourage over-selection of unnecessary regions in the mask.

\subsection{Network Parameters}
\label{sec:experiment_network_instantiation}

We instantiated our model (section~\ref{sec:approach}) with a predictive horizon of $H=12$ at 6 Hz timesteps for experiments involving planning low turbulence, collision-free trajectories in simulation (sections~\ref{sec:experiment_toy},~\ref{sec:experiment_smooth_trajectory_sim}), $H=12$ at $\approx{4}$ Hz for the corresponding real-world experiments (section~\ref{sec:experiment_smooth_trajectory_real}), and $H=16$ at 4 Hz for hill climbing in presence of low traction terrain (section~\ref{sec:experiment_icy_hill}). Adam~\cite{kingma2014adam} with a learning rate of $10^{-3}$ was used for optimization. L2 regularization with a factor of $1\times 10^{-4}$ was used for trajectory-attention and self-attention models while $5\times 10^{-4}$ was used for the baseline no-attention model as this resulted in better validation set performance. 

\subsection{Toy Experiment: Unreliability of Self-Attention}
\label{sec:experiment_toy}

\begin{figure}[t]
  \centering
  \includegraphics[width=0.95\linewidth]{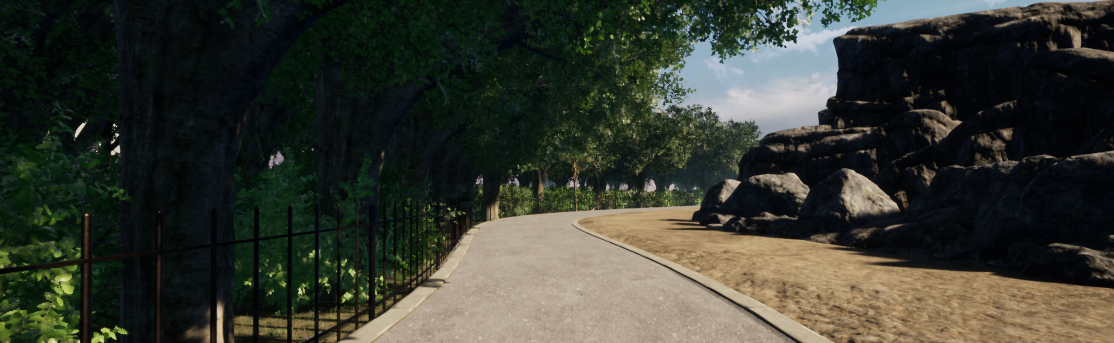}
  \caption{Toy environment for section~\ref{sec:experiment_toy} experiment consisting of two terrain types with correlated background scenery.}
\label{fig:experiment_toy_environment}
\end{figure}

\textbf{Environment and Data:} We begin with a motivating toy experiment to illustrate the potential unreliability of black-box self-attention methods. We simulated a short 150 meter outdoor trail in Unreal Engine~\cite{unrealengine} consisting of smooth and rough terrain classes (see Fig.~\ref{fig:experiment_toy_environment}). In this environment, terrain is correlated with background scenery, such that smooth terrain is associated with overhanging trees and foliage while rough terrain is identified by clear skies and proximity to rocky outcrops. We then made a test environment which swaps the position of the two terrain classes while the background was unchanged. A dataset of 14,000 samples was gathered from each environment in the form $(\vo_t, a_t, e_t^\mathrm{terr})$, defining the image observation, steering (throttle is assumed constant) and ground-truth terrain class of a categorical distribution. Data was gathered at 6 Hz using random off-policy exploration on a small-scale buggy moving at $\approx{6.9}$ m/s. The trajectory-attention and baseline models were trained with a 80/20 train/validation split (shuffled each random seeded trial) to predict terrain traversed over trajectories of $H=12$.

\begin{figure}[ht]
  \centering
  \includegraphics[width=1.0\linewidth]{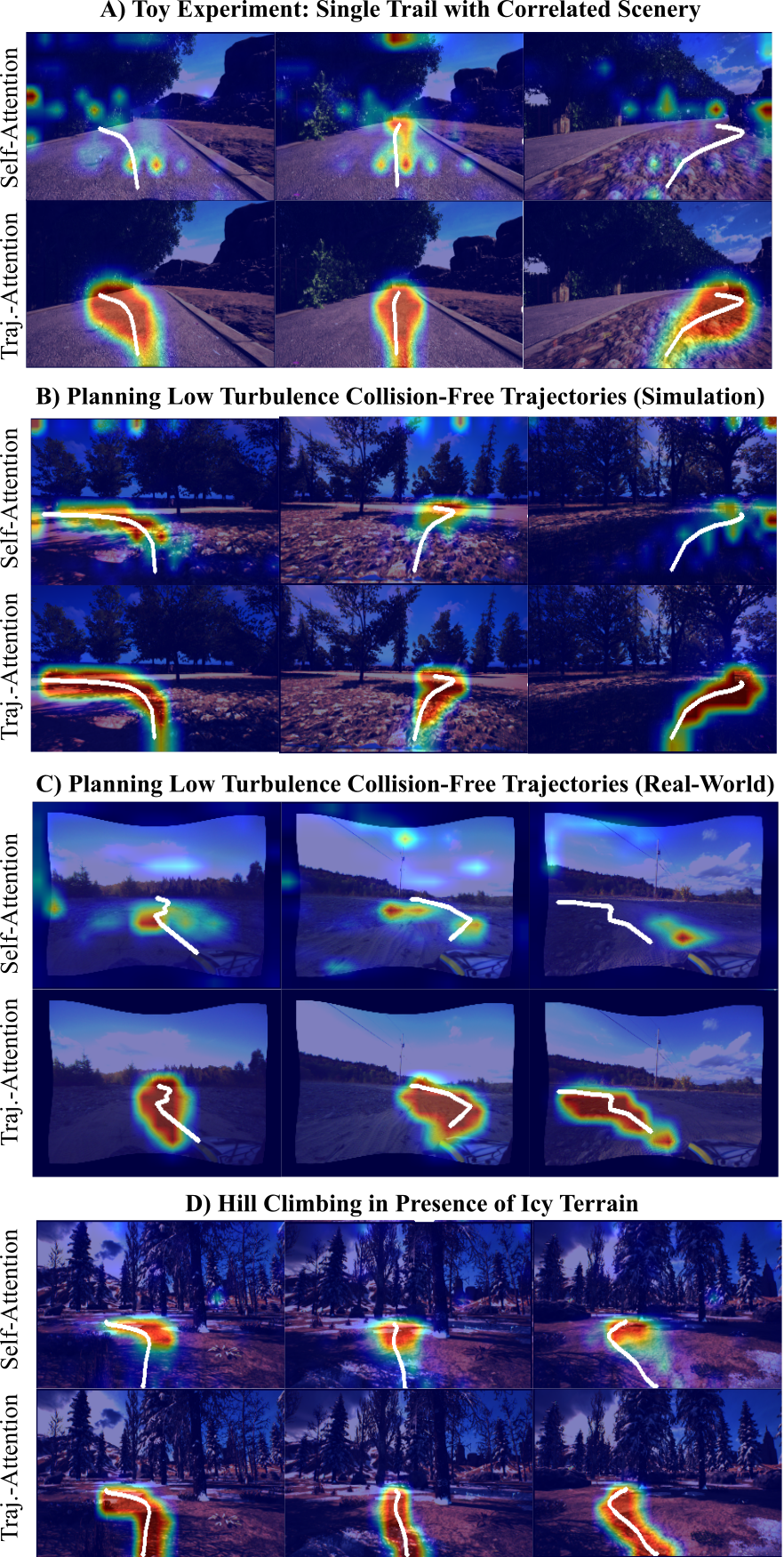}
  \vspace{-1em}
  \caption{Ground-truth trajectories (white) and predicted attention masks (superimposed over all timesteps) for various experiments comparing self-attention (top) and trajectory-constrained attention (bottom). The self-attention variant generally follows the trajectory but sometimes focuses on background features leading to diminished generalization performance.}
\label{fig:experiment_results_attention_masks}
\end{figure}

\textbf{Results and Discussion:} Intuitively, the trained model should learn to associate the visual terrain driven over with the ground-truth turbulence labels of this terrain. However, although the no-attention and self-attention variants performed well on the training environment's validation set, they performed poorly when generalizing to the test environment with swapped terrain positions (Table~\ref{tab:experiment_toy_predictive_accuracy}). Attention masks visualized in Fig.~\ref{fig:experiment_results_attention_masks} A) show that the self-attention network places significant attention on background scenery correlated with the terrain instead of the terrain itself. This issue is reduced with larger datasets of more randomized scenery that break misleading visual correlations between background scenery and the traversed terrain labels. Nonetheless, this result illustrates how black-box self-attention methods may learn unexpected or undesirable patterns if left unconstrained.

\setlength{\belowcaptionskip}{0pt}
\setlength\tabcolsep{8pt}
\begin{table}[h]
  \centering
  \caption{Predictive accuracy (averaged over trajectory timesteps and dataset) in toy environment (section~\ref{sec:experiment_toy}) over 6 random seeds ($\pm\sigma$). Test environment swaps terrain.} \label{tab:experiment_toy_predictive_accuracy}
  \vspace{0.0in}
    \begin{tabular}[t]{lcccc}
     \multicolumn{4}{c}{\textbf{Terrain Accuracy}} \\
     \hline
     \multicolumn{1}{l|}{Dataset} & No-Attention & Self-Attention & Traj.-Attention   \\
     \hline
         \multicolumn{1}{l|}{Validation}    & 93.40$\pm$0.91     & 94.23$\pm$1.00     & \textbf{95.00$\pm$0.46} \\
         \multicolumn{1}{l|}{Test}          & 46.16$\pm$4.44   & 58.06$\pm$7.13      &   \textbf{87.69$\pm$1.55} \\
     \hline
    \end{tabular}
\end{table}
\setlength{\belowcaptionskip}{-8pt}

\subsection{Planning Low Turbulence, Collision-Free Trajectories (Simulation)}
\label{sec:experiment_smooth_trajectory_sim}

\begin{figure}[t]
  \centering
  \includegraphics[width=1.0\linewidth]{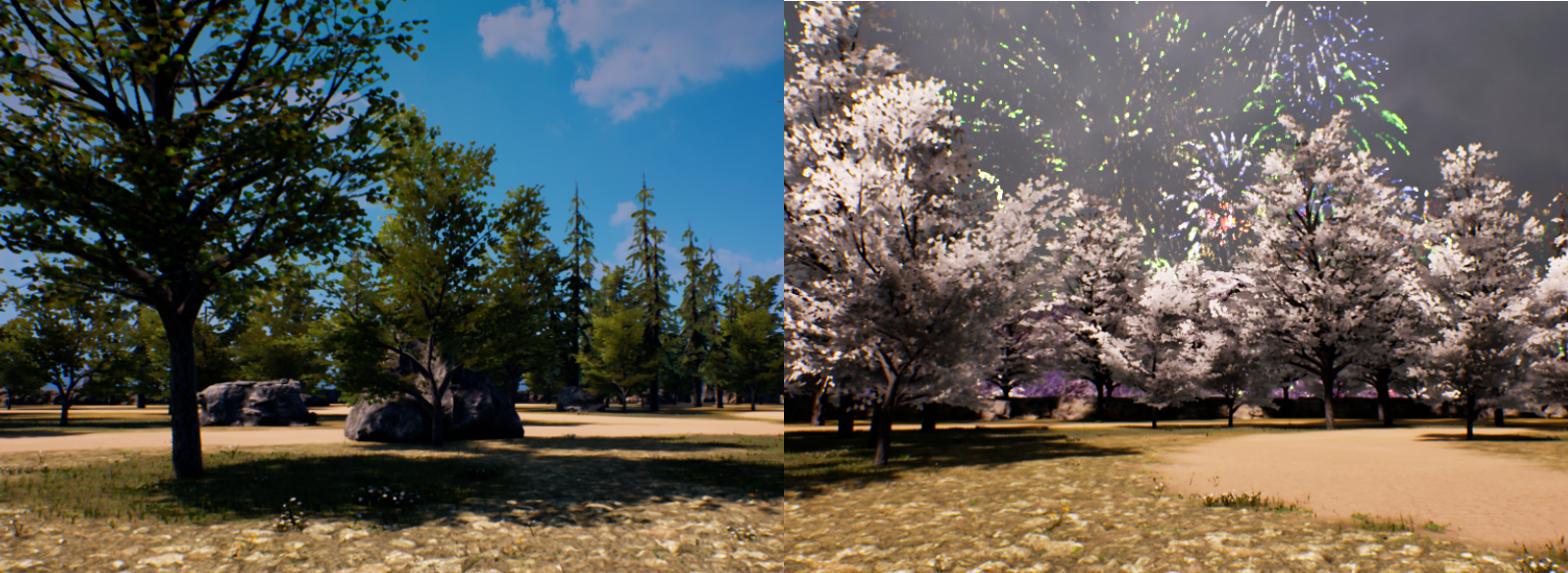}
  \caption{Training (left) and test (right) simulation environments for experiments in section \ref{sec:experiment_smooth_trajectory_sim}.}
\label{fig:simulation_environment}
\end{figure}

\textbf{Environment and Data:} We simulated a 190x190 meter procedural generated off-road environment in Unreal Engine \cite{unrealengine} with three terrain classes of increasing turbulence and various obstacles. In this environment, a small-scale buggy moving at $\approx{6.9}$ m/s was taught to plan smooth, collision-free trajectories. Visually distinctive training and test environments were used (see Fig.~\ref{fig:simulation_environment}). Training data was collected with random off-policy exploration at 6 Hz. Samples were labeled with the image observation $\vo_t$, steering action $a_t$, vehicle position $\ve^\mathrm{pos}_t$, terrain turbulence class $e^\mathrm{terr}_t$ and collision $e^\mathrm{coll}_t$ events. 60,000 samples were collected from the training environment to which random 80/20 train/validation set splits were done. An additional 60,000 test environment samples were collected to assess generalization performance. 

\textbf{Planner Instantiation:} After model learning, predicted trajectory events were transformed into rewards using:
\begin{equation}
R(\hat{\ve}_{t+1:t+H}^{0:K-1}) = -\sum_{i=1}^{H}{(1-\hat{e}_{t+i}^\mathrm{coll})\hat{e}_{t+i}^\mathrm{terr} + \hat{e}_{t+i}^\mathrm{coll}|E^\mathrm{terr}|}
\end{equation}
where $\hat{e}_{t+i}^\mathrm{terr}$ is the predicted turbulence class (higher values being more rough) within a set of size $|E^\mathrm{terr}|$ and $\hat{e}_{t+i}^\mathrm{coll}$ is the predicted collision probability. During planning, the reward function was optimized over actions using the \ac{CEM} planner (section \ref{sec:approach_planning}) to find a maximizing action rollout $\va_{t:t+H-1}^*$ from which the first action is set as the agent's next control and this process was repeated every timestep. We used $N=2048$ sampled rollouts and 2 iterations for the \ac{CEM} planner.

\begin{figure}[t]
  \centering
  \includegraphics[width=0.95\linewidth]{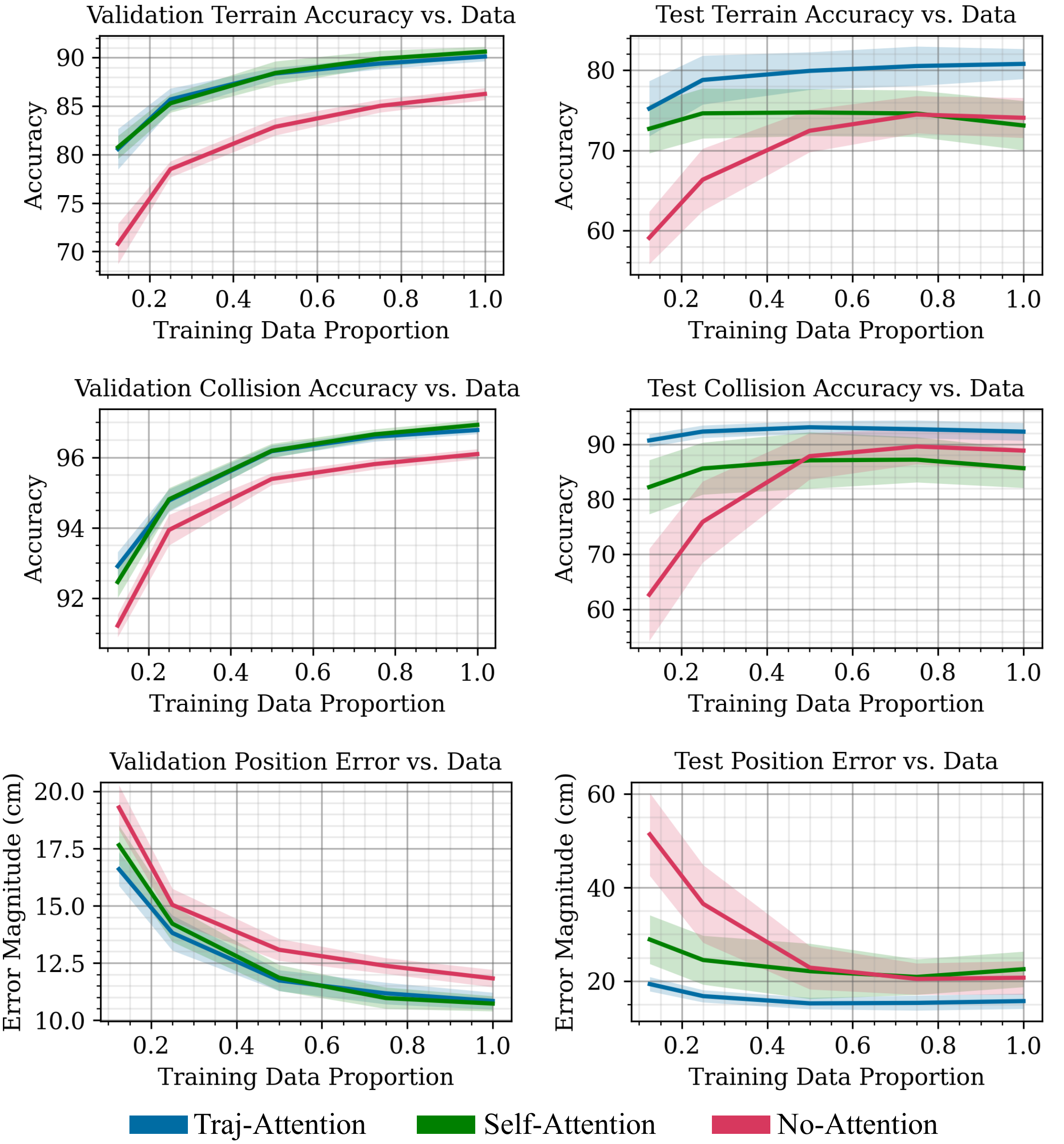}
  \caption{Average predictive accuracy over trajectory timesteps versus training dataset size on validation (left) and test (right) sets for part~\ref{sec:experiment_smooth_trajectory_sim} (planning low turbulence, collision-free trajectories) over 10 random seeds ($\pm\sigma$).}
\label{fig:experiment_results_smooth_traj_data_efficiency}
\end{figure}

\textbf{Results and Discussion:} Fig.~\ref{fig:experiment_results_smooth_traj_data_efficiency} shows the predictive accuracy on validation and test sets (repeated over several random trials, where each changes the train/validation split). Table~\ref{tab:experiment_results_onpolicy_smooth_traj} lists on-policy evaluation results obtained after model learning by running the resulting planner in the training and test environments. For on-policy evaluation we ran 90 short 30-second episodes ($\approx 210$ meters) at randomized starting locations (though shared by all models for fairness).

The attention models performed similarly on the static validation set and during on-policy evaluation in the training environment. However the trajectory-constrained attention variant generalized best to the test environment. These results appear to indicate that the no-attention and self-attention models overfit on background features which change significantly in the test environment, leading to reduced test performance. The self-attention and trajectory-constrained attention masks are compared in Fig.~\ref{fig:experiment_results_attention_masks} B). The predicted attention positions of the self-attention variant generally followed the ground-truth trajectory but in certain instances placed attention on background features.
We also found that our model possessed improved data efficiency (learning faster with fewer samples) as shown in the predictive accuracy results versus dataset size plotted in Fig.~\ref{fig:experiment_results_smooth_traj_data_efficiency}. 

\setlength{\belowcaptionskip}{0pt}
\setlength\tabcolsep{8pt}
\begin{table}[b]
  \centering
  \caption{On-policy evaluation for 90 random start position episodes ($\pm\sigma$) for planning low turbulence, collision-free trajectories (simulation). Episodic return derived from sum of traversed terrain (+3 for smooth to +1 for rough). } \label{tab:experiment_results_onpolicy_smooth_traj}
  \vspace{0.0in}
    \begin{tabular}[t]{lcccc}

     \multicolumn{4}{c}{\textbf{Episodic Return}} \\
     \hline
     \multicolumn{1}{l|}{Environment} & No-Attention & Self-Attention & Traj.-Attention   \\
     \hline
         \multicolumn{1}{l|}{Train}    & 482.1$\pm$89.6     & 502.5$\pm$61.2     & \textbf{507.5$\pm$72.5} \\
         \multicolumn{1}{l|}{Test}          & 380.9$\pm$101.6   & 396.0$\pm$120.6      &   \textbf{444.8$\pm$113.1} \\
     \hline
     \\

     \multicolumn{4}{c}{\textbf{Percent Episode Completed Before Collision}} \\
     \hline
     \multicolumn{1}{l|}{Environment} & No-Attention & Self-Attention & Traj.-Attention   \\
     \hline
         \multicolumn{1}{l|}{Train}    & 96.9$\pm$12.4     & \textbf{99.5$\pm$5.9}     & 98.4$\pm$11.8 \\
         \multicolumn{1}{l|}{Test}          & 87.3$\pm$21.0   & 87.9$\pm$24.6      &   \textbf{90.8$\pm$19.2} \\
     \hline

    \end{tabular}
\end{table}
\setlength{\belowcaptionskip}{-8pt}

\subsection{Planning Low Turbulence, Collision-Free Trajectories (Real-World)}
\label{sec:experiment_smooth_trajectory_real}

\begin{figure}[t]
  \centering
  \includegraphics[width=1.0\linewidth]{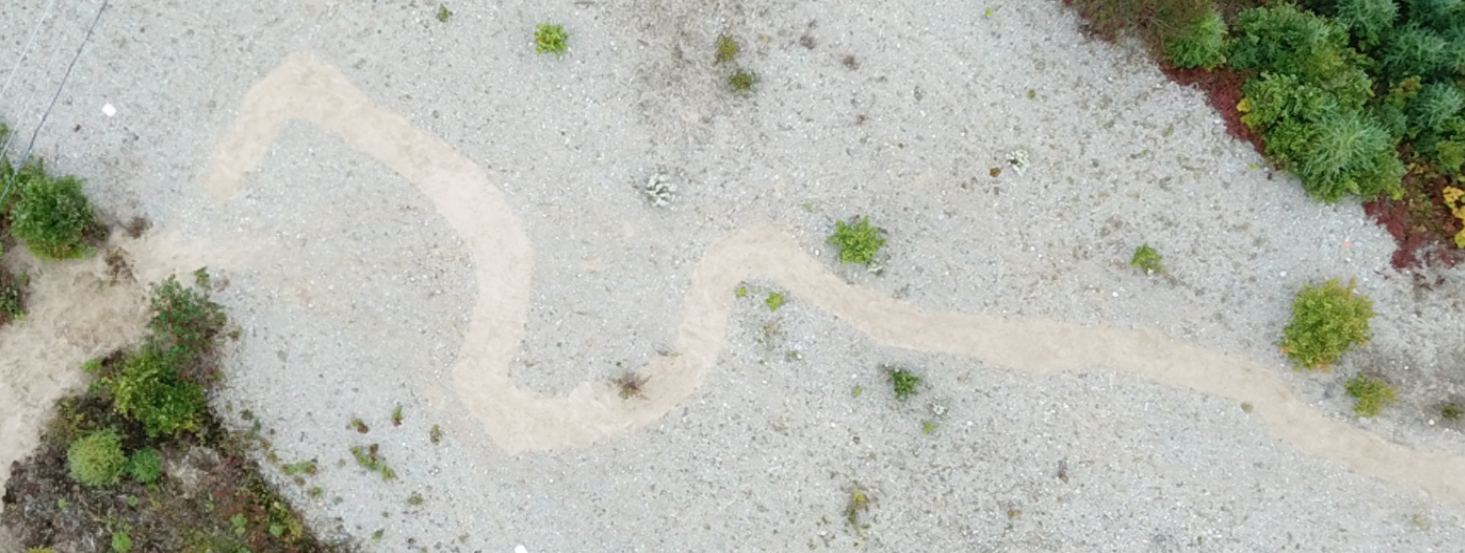}
  \caption{Real-world environment for section ~\ref{sec:experiment_smooth_trajectory_real} experiment. Environment consists two terrain classes (smooth and rough) with foliage obstacles.}
\label{fig:experiment_real_world_smooth_traj_environment}
\end{figure}

\textbf{Environment and Data:} We repeated the experimental evaluation of section~\ref{sec:experiment_smooth_trajectory_sim} using a real-world dataset~\cite{air_ground_planning} consisting of 12,000 labelled samples specifying image observation, steering action, binary terrain class (rough or smooth), collision and pose information at each timestep. Data was collected by random exploration of a 1:5 scale electric remote-controlled buggy moving at speeds of $\approx1.6$ m/s in an off-road environment spanning roughly 0.25~$\mathrm{km}^2$ (see Fig.~\ref{fig:experiment_real_world_smooth_traj_environment}). Vehicle poses and roughness were labeled with high precision~\ac{RTK}~\ac{GPS} and aerial segmentation while collisions were labeled by LIDAR proximity. Because no separate test environment was available, only a validation set was used to evaluate performance, where a 80/20 train/validation set split was randomly done over repeated trials. 

\begin{figure}[t]
  \centering
  \includegraphics[width=1.0\linewidth]{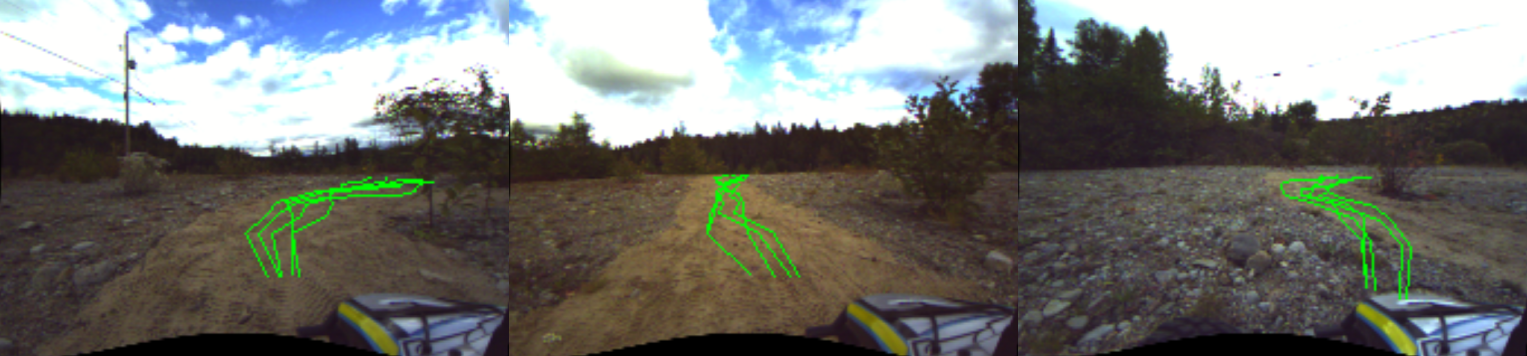}
  \caption{High-scoring action rollouts moving towards smooth terrain output by the trajectory-attention planner when invoked on sample image inputs.}
\label{fig:experiment_real_world_smooth_traj_sample_paths}
\end{figure}

\begin{figure*}[t!]
  \centering
  \includegraphics[width=1.0\linewidth]{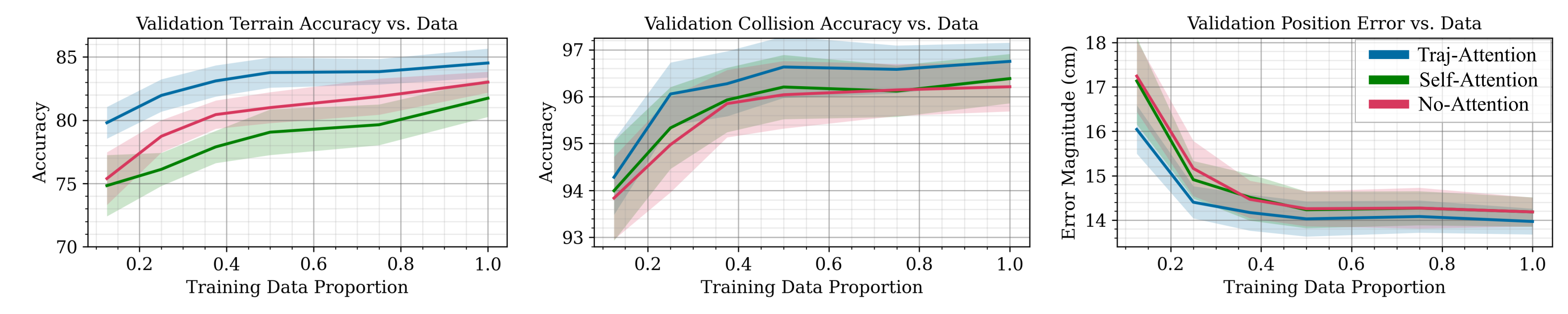}
  \caption{Average predictive accuracy over trajectory timesteps versus training dataset size on evaluation set for experiment~\ref{sec:experiment_smooth_trajectory_real} (planning low turbulence, collision-free trajectories (real-world)) over 6 random seeded trials ($\pm\sigma$). }
\label{fig:experiment_rw_pred_accuracy}
\end{figure*}

\textbf{Results and Discussion:} Fig.~\ref{fig:experiment_rw_pred_accuracy} plots predictive accuracy on the evaluation set while Fig.~\ref{fig:experiment_results_attention_masks} C) compares masks of the attention models. Some improvement was found in terrain classification while the difference in other metrics was mostly insignificant. Similar to simulation results, we found that the self-attention model placed some attention on background features resulting in worse generalization.  Fig.~\ref{fig:experiment_real_world_smooth_traj_sample_paths} shows high-scoring action sequences output by the planner.


\subsection{Hill Climbing in Presence of Slippery Terrain}
\label{sec:experiment_icy_hill}

\begin{figure}[t]
  \centering
  \includegraphics[width=0.95\linewidth]{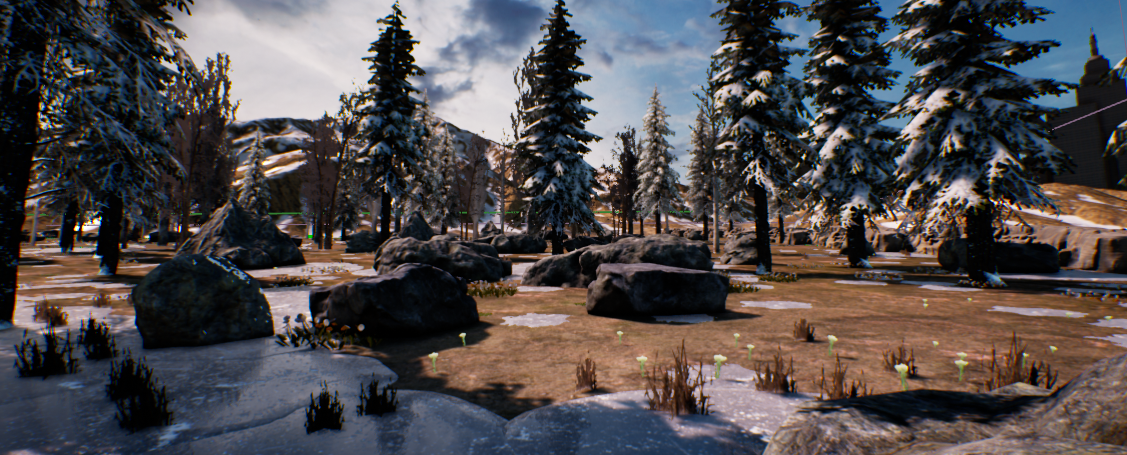}
  \caption{Simulated icy hill training environment for section \ref{sec:experiment_icy_hill} experiment.}
\label{fig:experiment_icy_hill_environment}
\end{figure}

\begin{figure}[t]
  \centering
  \includegraphics[width=1.0\linewidth]{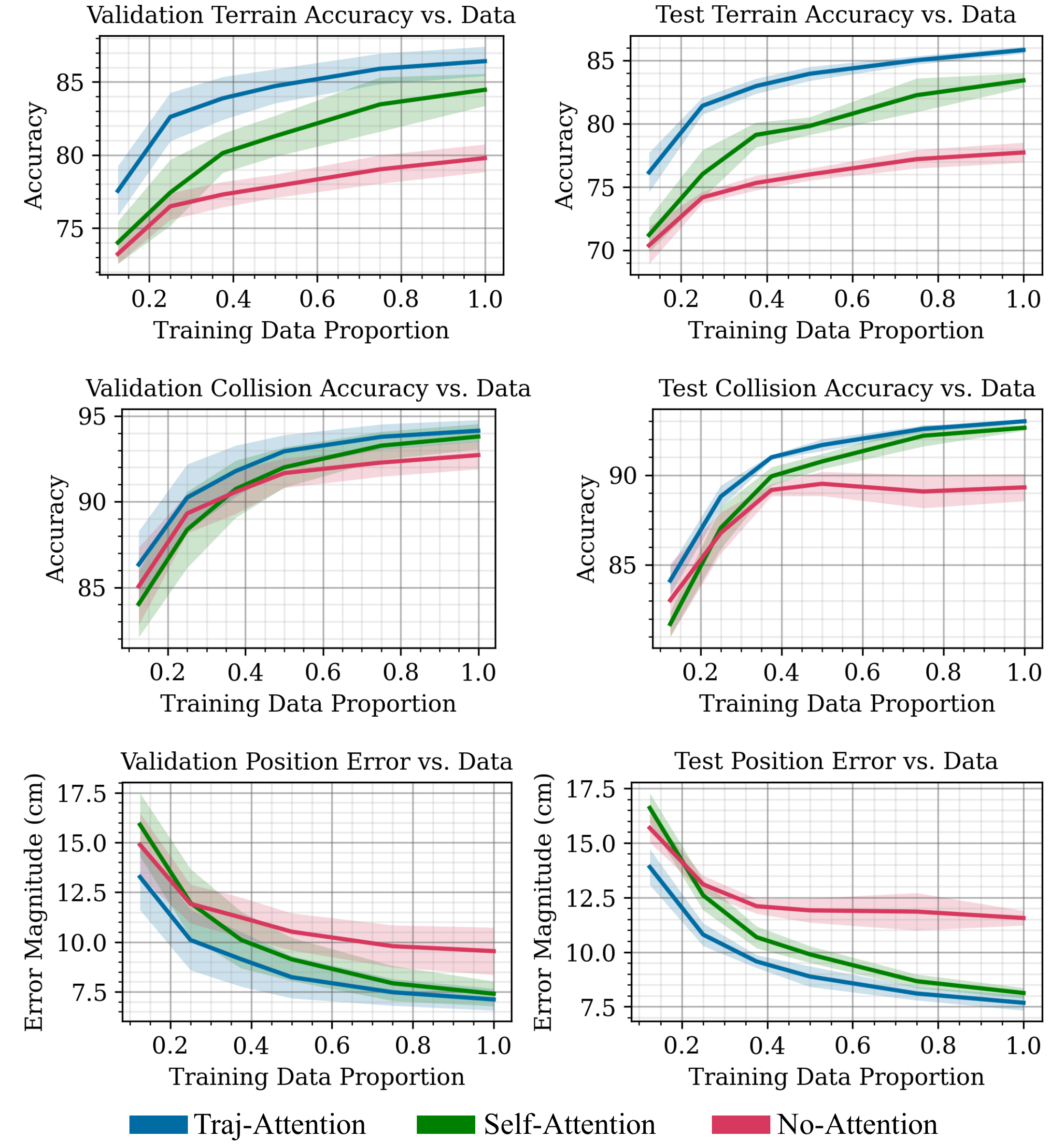}
    \caption{Average predictive accuracy over trajectory timesteps versus training dataset size on validation (left) and test (right) sets for experiment~\ref{sec:experiment_icy_hill} (hill climbing in presence of slippery terrain) over 6 random seeds ($\pm\sigma$).}
\label{fig:experiment_results_icy_hill_data_efficiency}
\end{figure}

We applied our method to learning a navigation policy for driving up a hill in the presence of slippery terrain. In this environment, selectable locking differentials were added to the action space. Locking differentials are a mechanism equipped on some off-road vehicles used to overcome sections of low traction terrain, but reduce turning ability and increase tire wear, so should only be used when required.

\textbf{Environment and Data}: A randomized procedural generated 190x190 meter ice-covered hill was made in simulation (see Fig.~\ref{fig:experiment_icy_hill_environment}). Our learner was a small-scale automated buggy capable of $\approx{2.5}$ m/s speeds. The environment was composed of two terrain types (ice and dirt) with various trees and large rocks acting as obstacles. Three randomized test environments were made with different arrangements of terrain, obstacles and scenery. We collected 72,000 training and 108,000 test samples using random off-policy exploration at 4 Hz, specifying the image observation $\vo_t$, steering and binary differential lock action $\va_t$, terrain class $e^\mathrm{terr}_t$, collision/stuck event $e^\mathrm{coll}_t$ and vehicle pose $\ve^\mathrm{pos}_t$.

\textbf{Network Instantiation:} In this environment we plan over locking differentials in addition to steering. The joint planning of these actions over the full trajectory length $H$ increases planning time significantly. Instead we trained two separate control networks (Fig. \ref{fig:network_architecture}): one for short-term planning over locking differentials ($H=2$) and one for long-term planning over steering actions ($H=16$). The differential locking network was trained to predict tire slip events conditioned on steering actions and rewarded to only lock when slipping. The steering control network was then trained using data collected with an expert locking policy running in the background. 
We also found it beneficial to give context of what terrain the vehicle is currently on. To this end we fed the last 3 historic images into an additional \ac{RNN} with 256 neuron cells and using the same image encoder as Fig.~\ref{fig:network_architecture} with optional attention on these historic images depending on the model. In the self-attention model the attention mask is learned while for the trajectory-attention model the vehicle's current location is selected since it is known for the current timestep. Although this selection may seem as a strict advantage for our method, it should be noted that the baselines have the flexibility to use this historic context to enhance future predictions beyond only receiving information of the vehicle’s current terrain. The final output hidden state of this historic network is concatenated to the initial hidden state of the \ac{RNN}s in Fig.~\ref{fig:network_architecture}.

\textbf{Planner Instantiation}: Planning was done in two stages. Rewards predicted by the steering network (Eq.~\ref{eq:icy_hill_steer_reward}) were first used to optimize for an action sequence $\va^*_{t:t+H-1}$ (following section~\ref{sec:approach_planning}) to move towards the hill's top as quickly as possible while avoiding collisions and stuck events.
\begin{equation}\label{eq:icy_hill_steer_reward}
R(\hat{\ve}_{t+1:t+H}^{0:K-1}) = \sum_{i=1}^{H}{(1-\hat{e}_{t+i}^\mathrm{coll})(\hat{\ve}^{\Delta\mathrm{pos}}_{t+i} \cdot \vg)-\hat{e}_{t+i}^\mathrm{coll}}
\end{equation}
where $\hat{\ve}^{\Delta\mathrm{pos}}_{i+1}$ is the predicted change in position per timestep, $\vg$ is a unit goal vector to the hill top (both in the robot's coordinate frame) and $\hat{e}_{t+i}^\mathrm{coll}$ is the probability of collision. 

The differential locking network then plans over locking actions to eliminate predicted tire slip events over the short $H=2$ horizon, given the computed steering actions as inputs.

\setlength{\belowcaptionskip}{0pt}
\setlength\tabcolsep{5pt}
\begin{table}[h]
  \centering
  \caption{On-policy evaluation for hill climbing experiment over 90 training and 270 test environment episodes ($\pm\sigma$) starting at random locations along the hill base for planner trained with 100\% and 50\% of data. } \label{tab:experiment_results_icy_hill_on_policy_evaluation}
  \vspace{-0.05in}
    \begin{tabular}[t]{lcccc}
     
      \multicolumn{4}{c}{\textbf{Percent Route To Goal Completed}} \\
     \hline
     \multicolumn{1}{l|}{Environment} & No-Attention & Self-Attention & Traj.-Attention   \\
     \hline
     \multicolumn{1}{l|}{Train (100\% data)}    & 72.1$\pm$35.4     & 74.9$\pm$33.6     & \textbf{80.1$\pm$31.0} \\
     \multicolumn{1}{l|}{Test (100\% data)}          & 34.6$\pm$27.8   & 64.2$\pm$34.1      &   \textbf{65.4$\pm$34.5} \\
     \hline
         \multicolumn{1}{l|}{Train (50\% data)}    & 50.5$\pm$33.0     & 62.9$\pm$30.9     & \textbf{70.3$\pm$32.7} \\
         \multicolumn{1}{l|}{Test (50\% data)}          & 28.3$\pm$26.1   & 38.0$\pm$29.7      &   \textbf{58.3$\pm$33.8} \\
     \hline
     \\

     \multicolumn{4}{c}{\textbf{Rate to Goal (m/s)}} \\
     \hline
     \multicolumn{1}{l|}{Environment} & No-Attention & Self-Attention & Traj.-Attention   \\
     \hline
     \multicolumn{1}{l|}{Train (100\% data)}    & 1.80$\pm$0.26     & \textbf{1.82$\pm$0.21}     & \textbf{1.82$\pm$0.20} \\
     \multicolumn{1}{l|}{Test (100\% data)}          & 1.43$\pm$0.32   & 1.63$\pm$0.25      &   \textbf{1.64$\pm$0.28} \\
     \hline
         \multicolumn{1}{l|}{Train (50\% data)}    & 1.64$\pm$0.29     & 1.69$\pm$0.18     & \textbf{1.72$\pm$0.23} \\
         \multicolumn{1}{l|}{Test (50\% data)}          & 1.36$\pm$0.32   & 1.39$\pm$0.29      &   \textbf{1.52$\pm$0.31} \\
     \hline
     
    \end{tabular}
\end{table}
\setlength{\belowcaptionskip}{-8pt}

\textbf{Results and Discussion:} Fig.~\ref{fig:experiment_results_icy_hill_data_efficiency} shows the predictive accuracy after training while Table~\ref{tab:experiment_results_icy_hill_on_policy_evaluation} lists on-policy results. The trajectory-attention model performed slightly better in the limit of training data, but more significant gains occurred at smaller dataset sizes, indicating greater data efficiency. Fig.~\ref{fig:experiment_results_attention_masks} D) compares the predicted masks of the attention models. Opposed to previous experiment, the self-attention model more reliably followed vehicle trajectories, perhaps due to the larger dataset of more dense random scenery causing more invariance to irrelevant background features.

\section{Conclusion} 
\label{sec:conclusion}
In this work we have presented a reward-predictive, model-based learning method augmented with trajectory-constrained visual attention to enhance predictive accuracy during local planning. Our method was validated in visual navigation tasks of planning low terrain turbulence, collision-free trajectories and hill climbing in the presence of slippery terrain. We found improved generalization and data efficiency compared to no-attention and self-attention baselines.

\section{Acknowledgements} 
\label{sec:acknowledgements}
This work was supported by CSA STDP 6.1 (Canadian Space Agency Space Technology Development Program), NCRN (NSERC Canadian Robotics Network) and consultation with CM Labs~\cite{cmlabs}.

\begin{acronym}
\acro{CNN}{Convolutional Neural Network}
\acro{RNN}{Recurrent Neural Network}
\acro{VPN}{Value Prediction Network}
\acro{GCG}{General Computation Graph}
\acro{DNN}{Deep Neural Network}
\acro{FOV}{Field of View}
\acro{ORB}{Oriented FAST and Rotated BRIEF}
\acro{DSO}{Direct Sparse Odometry}
\acro{MAV}{Micro Air Vehicle}
\acro{UAV}{Unmanned Aerial Vehicle}
\acro{IMU}{Inertial Measurement Unit}
\acro{ROS}{Robot Operating System}
\acro{GPS}{Global Positioning System}
\acro{RTK}{Real-Time Kinematic}
\acro{RMS}{Root Mean Square}
\acro{SVM}{Support Vector Machine}
\acro{LSTM}{Long Short-Term Memory}
\acro{MPC}{Model Predictive Control}
\acro{MLE}{Maximum Likelihood Estimation}
\acro{CEM}{Cross Entropy Method}
\acro{NLP}{Natural Language Processing}
\end{acronym}



\bibliographystyle{IEEEtran}
\bibliography{MyBSTcontrol, custom, library}

\begin{thebibliography}{10}
\providecommand{\url}[1]{#1}
\csname url@samestyle\endcsname
\providecommand{\newblock}{\relax}
\providecommand{\bibinfo}[2]{#2}
\providecommand{\BIBentrySTDinterwordspacing}{\spaceskip=0pt\relax}
\providecommand{\BIBentryALTinterwordstretchfactor}{4}
\providecommand{\BIBentryALTinterwordspacing}{\spaceskip=\fontdimen2\font plus
\BIBentryALTinterwordstretchfactor\fontdimen3\font minus
  \fontdimen4\font\relax}
\providecommand{\BIBforeignlanguage}[2]{{%
\expandafter\ifx\csname l@#1\endcsname\relax
\typeout{** WARNING: IEEEtran.bst: No hyphenation pattern has been}%
\typeout{** loaded for the language `#1'. Using the pattern for}%
\typeout{** the default language instead.}%
\else
\language=\csname l@#1\endcsname
\fi
#2}}
\providecommand{\BIBdecl}{\relax}
\BIBdecl

\bibitem{Giusti2016}
A.~Giusti, J.~Guzzi, D.~C. Cireşan, F.~L. He, J.~P. Rodr{\'{i}}guez,
  F.~Fontana, M.~Faessler, C.~Forster, J.~Schmidhuber, G.~D. Caro,
  D.~Scaramuzza, and L.~M. Gambardella, ``{A Machine Learning Approach to
  Visual Perception of Forest Trails for Mobile Robots},'' \emph{IEEE Robotics
  and Automation Letters}, vol.~1, no.~2, pp. 661--667, July 2016.

\bibitem{Smolyanskiy2017}
N.~Smolyanskiy, A.~Kamenev, J.~Smith, and S.~Birchfield, ``{Toward low-flying
  autonomous MAV trail navigation using deep neural networks for environmental
  awareness},'' in \emph{2017 IEEE/RSJ International Conference on Intelligent
  Robots and Systems (IROS)}, Sept 2017, pp. 4241--4247.

\bibitem{bojarski2016end}
M.~Bojarski, D.~{Del Testa}, D.~Dworakowski, B.~Firner, B.~Flepp, P.~Goyal,
  L.~D. Jackel, M.~Monfort, U.~Muller, J.~Zhang, and Others, ``{End to end
  learning for self-driving cars},'' \emph{arXiv preprint arXiv:1604.07316},
  2016.

\bibitem{Wellhausen2019}
L.~{Wellhausen}, A.~{Dosovitskiy}, R.~{Ranftl}, K.~{Walas}, C.~{Cadena}, and
  M.~{Hutter}, ``{Where Should I Walk? Predicting Terrain Properties From
  Images Via Self-Supervised Learning},'' \emph{IEEE Robotics and Automation
  Letters}, vol.~4, no.~2, pp. 1509--1516, April 2019.

\bibitem{kahn2020badgr}
G.~Kahn, P.~Abbeel, and S.~Levine, ``{BADGR}: An autonomous self-supervised
  learning-based navigation system,'' \emph{arXiv preprint arXiv:2002.05700},
  2020.

\bibitem{air_ground_planning}
T.~{Manderson}, S.~{Wapnick}, D.~{Meger}, and G.~{Dudek}, ``Learning to drive
  off road on smooth terrain in unstructured environments using an on-board
  camera and sparse aerial images,'' in \emph{2020 IEEE International
  Conference on Robotics and Automation}, 2020, pp. 1263--1269.

\bibitem{iyer_transparency_2018}
R.~Iyer, Y.~Li, H.~Li, M.~Lewis, R.~Sundar, and K.~Sycara, ``Transparency and
  explanation in deep reinforcement learning neural networks,'' in
  \emph{Proceedings of the 2018 AAAI/ACM Conference on AI, Ethics, and
  Society}, 2018, p. 144–150.

\bibitem{NIPS2018_outofdistribution}
K.~Lee, K.~Lee, H.~Lee, and J.~Shin, ``A simple unified framework for detecting
  out-of-distribution samples and adversarial attacks,'' in \emph{Advances in
  Neural Information Processing Systems 31}.\hskip 1em plus 0.5em minus
  0.4em\relax Curran Associates, Inc., 2018, pp. 7167--7177.

\bibitem{deep_learning_book}
I.~Goodfellow, Y.~Bengio, and A.~Courville, \emph{Deep Learning}.\hskip 1em
  plus 0.5em minus 0.4em\relax MIT Press, 2016.

\bibitem{jetley2018_learn_to_pay_attention}
S.~Jetley, N.~A. Lord, N.~Lee, and P.~Torr, ``Learn to pay attention,'' in
  \emph{International Conference on Learning Representations}, 2018.

\bibitem{Schlemper2019_attentiongatednetworks}
J.~Schlemper, O.~Oktay, M.~Schaap, M.~Heinrich, B.~Kainz, B.~Glocker, and
  D.~Rueckert, ``Attention gated networks: Learning to leverage salient regions
  in medical images,'' in \emph{Medical Image Analysis, Volume 53}, 2019, pp.
  197 -- 207.

\bibitem{vaswani2017_transformers}
A.~Vaswani, N.~Shazeer, N.~Parmar, J.~Uszkoreit, L.~Jones, A.~N. Gomez, L.~u.
  Kaiser, and I.~Polosukhin, ``Attention is all you need,'' in \emph{Advances
  in Neural Information Processing Systems}, vol.~30.\hskip 1em plus 0.5em
  minus 0.4em\relax Curran Associates, Inc., 2017, pp. 5998--6008.

\bibitem{pomerleau1989alvinn}
D.~A. Pomerleau, ``{Alvinn: An autonomous land vehicle in a neural network},''
  in \emph{Advances in neural information processing systems}, 1988, pp.
  305--313.

\bibitem{ross2011reduction}
S.~Ross, G.~Gordon, and D.~Bagnell, ``{A reduction of imitation learning and
  structured prediction to no-regret online learning},'' in \emph{Proceedings
  of the fourteenth international conference on artificial intelligence and
  statistics}, 2011, pp. 627--635.

\bibitem{ensemble_dagger_2019}
K.~{Menda}, K.~{Driggs-Campbell}, and M.~J. {Kochenderfer}, ``Ensemble{DA}gger:
  A bayesian approach to safe imitation learning,'' in \emph{2019 IEEE/RSJ
  International Conference on Intelligent Robots and Systems (IROS)}, 2019, pp.
  5041--5048.

\bibitem{hg_dagger_2019}
M.~{Kelly}, C.~{Sidrane}, K.~{Driggs-Campbell}, and M.~J. {Kochenderfer},
  ``H{G}-{DA}gger: Interactive imitation learning with human experts,'' in
  \emph{2019 International Conference on Robotics and Automation (ICRA)}, 2019,
  pp. 8077--8083.

\bibitem{lee_sensor_fusion_uncertainty_2019}
K.~{Lee}, Z.~{Wang}, B.~{Vlahov}, H.~{Brar}, and E.~A. {Theodorou}, ``Ensemble
  bayesian decision making with redundant deep perceptual control policies,''
  in \emph{2019 18th IEEE International Conference On Machine Learning And
  Applications (ICMLA)}, 2019, pp. 831--837.

\bibitem{Oh2017}
J.~Oh, S.~Singh, and H.~Lee, ``Value prediction network,'' in \emph{Proceedings
  of the 31st International Conference on Neural Information Processing
  Systems}.\hskip 1em plus 0.5em minus 0.4em\relax Curran Associates Inc.,
  2017, pp. 6120--6130.

\bibitem{hafner2018planet}
D.~Hafner, T.~Lillicrap, I.~Fischer, R.~Villegas, D.~Ha, H.~Lee, and
  J.~Davidson, ``Learning latent dynamics for planning from pixels,'' in
  \emph{Proceedings of the 36th International Conference on Machine Learning},
  vol.~97, 09--15 June 2019, pp. 2555--2565.

\bibitem{hafner2020dreamer}
D.~Hafner, T.~Lillicrap, J.~Ba, and M.~Norouzi, ``Dream to control: Learning
  behaviors by latent imagination,'' in \emph{Eighth International Conference
  on Learning Representations}.\hskip 1em plus 0.5em minus 0.4em\relax {ICLR},
  2020.

\bibitem{hafner2021mastering}
D.~Hafner, T.~P. Lillicrap, M.~Norouzi, and J.~Ba, ``Mastering {A}tari with
  discrete world models,'' in \emph{International Conference on Learning
  Representations}.\hskip 1em plus 0.5em minus 0.4em\relax {ICLR}, 2021.

\bibitem{Kahn2018}
G.~Kahn, A.~Villaflor, B.~Ding, P.~Abbeel, and S.~Levine, ``{Self-Supervised
  Deep Reinforcement Learning with Generalized Computation Graphs for Robot
  Navigation},'' in \emph{Proceedings - IEEE International Conference on
  Robotics and Automation}, 2018, pp. 5129--5136.

\bibitem{RAM_VisualAttention}
V.~Mnih, N.~Heess, A.~Graves, and K.~Kavukcuoglu, ``Recurrent models of visual
  attention,'' in \emph{Advances in Neural Information Processing Systems
  27}.\hskip 1em plus 0.5em minus 0.4em\relax Curran Associates, Inc., 2014,
  pp. 2204--2212.

\bibitem{pmlr-v37-xuc15}
K.~Xu, J.~Ba, R.~Kiros, K.~Cho, A.~Courville, R.~Salakhudinov, R.~Zemel, and
  Y.~Bengio, ``Show, attend and tell: Neural image caption generation with
  visual attention,'' in \emph{Proceedings of the 32nd International Conference
  on Machine Learning}, vol.~37, 07--09 Jul 2015, pp. 2048--2057.

\bibitem{pmlr-v80-parmar18a_image_transformer}
N.~Parmar, A.~Vaswani, J.~Uszkoreit, L.~Kaiser, N.~Shazeer, A.~Ku, and D.~Tran,
  ``Image transformer,'' in \emph{Proceedings of the 35th International
  Conference on Machine Learning}, J.~Dy and A.~Krause, Eds., vol.~80.\hskip
  1em plus 0.5em minus 0.4em\relax PMLR, 10--15 Jul 2018, pp. 4055--4064.

\bibitem{carion2020endtoendtransformers}
N.~Carion, F.~Massa, G.~Synnaeve, N.~Usunier, A.~Kirillov, and S.~Zagoruyko,
  ``End-to-end object detection with transformers,'' in \emph{European
  Conference on Computer Vision (ECCV)}, 2020.

\bibitem{lee2019_perceptualattncontrol}
K.~Lee, G.~N. An, V.~Zakharov, and E.~A. Theodorou, ``Perceptual
  attention-based predictive control,'' in \emph{3rd Annual Conference on Robot
  Learning}, vol. 100.\hskip 1em plus 0.5em minus 0.4em\relax {PMLR}, 2019, pp.
  220--232.

\bibitem{drews2019}
P.~Drews, ``Visual attention for high speed driving,'' Ph.D. dissertation,
  Georgia Institute of Technology, 2019.

\bibitem{Zhang_2018_ECCV}
R.~Zhang, Z.~Liu, L.~Zhang, J.~A. Whritner, K.~S. Muller, M.~M. Hayhoe, and
  D.~H. Ballard, ``A{GIL}: Learning attention from human for visuomotor
  tasks,'' in \emph{Proceedings of the European Conference on Computer Vision
  (ECCV)}, September 2018.

\bibitem{liu2021gaze}
C.~Liu, Y.~Chen, M.~Liu, and B.~E. Shi, ``Using eye gaze to enhance
  generalization of imitation networks to unseen environments,'' \emph{IEEE
  Transactions on Neural Networks and Learning Systems}, vol.~32, no.~5, pp.
  2066--2074, 2021.

\bibitem{Hochreiter1997}
S.~Hochreiter and J.~Schmidhuber, ``Long short-term memory,'' \emph{Neural
  Comput.}, vol.~9, no.~8, pp. 1735--1780, Nov. 1997.

\bibitem{cem_method}
R.~Rubinstein, ``The cross-entropy method for combinatorial and continuous
  optimization,'' \emph{Methodology And Computing In Applied Probability},
  vol.~1, pp. 127--190, 1999.

\bibitem{kingma2014adam}
D.~P. Kingma and J.~Ba, ``Adam: {A} method for stochastic optimization,'' in
  \emph{3rd International Conference on Learning Representations}, 2015.

\bibitem{unrealengine}
{Epic Games}, ``Unreal engine,'' \url{https://www.unrealengine.com}.

\bibitem{cmlabs}
{CM Labs}, ``Cm labs,'' \url{https://www.cm-labs.com/}.

\end{thebibliography}

\maxpage[Total Document is too long. IROS limit is 6 pages plus 2 more for cash.]{8}

\end{document}